\documentclass[final]{cvpr}

\usepackage{times}
\usepackage{epsfig}
\usepackage{graphicx}
\usepackage{amsmath}
\usepackage{amssymb}
\usepackage{color}
\usepackage{booktabs}
\usepackage{multirow}
\usepackage[pagebackref=true, breaklinks=true, colorlinks, bookmarks=false]{hyperref}

\def\etal{\emph{et al.}}
\def\ie{\emph{i.e.}}

\def\confYear{CVPR 2021}

\begin{document}

\title{Region-Adaptive Deformable Network for Image Quality Assessment}

\author{
Shuwei Shi$^{1}$\textsuperscript{\thanks{The first two authors contribute equally to this work. }}  
\quad Qingyan Bai$^{1^{*}}$
\quad Mingdeng Cao$^{2}$ 
\quad Weihao Xia$^{1}$ \quad \\ 
Jiahao Wang$^{2}$ 
\quad Yifan Chen$^{1}$ 
\quad Yujiu Yang$^{1}\thanks{This work was supported partially the Major Research Plan of National Natural Science Foundation of China (Grant No. 61991450) and the Shenzhen Science and Technology Project under Grant (ZDYBH201900000002, JCYJ20180508152042002).}$\\
	$^{1}$Tsinghua Shenzhen International Graduate School, Tsinghua University\\
	$^{2}$Department of Automation, Tsinghua University\\
	{\tt\small \{ssw20, bqy20, cmd19, wang-jh19, chenyf20\}@mails.tsinghua.edu.cn}  \\
	{\tt\small xiawh3@outlook.com}  \quad{\tt\small yang.yujiu@sz.tsinghua.edu.cn} 
}
\maketitle

\pagestyle{empty}  %
\thispagestyle{empty} %

\begin{abstract}
Image quality assessment (IQA) aims to assess the perceptual quality of images. The outputs of the IQA algorithms are expected to be consistent with human subjective perception. In image restoration and enhancement tasks, images generated by generative adversarial networks (GAN) can achieve better visual performance than traditional CNN-generated images, although they have spatial shift and texture noise. Unfortunately, the existing IQA methods have unsatisfactory performance on the GAN-based distortion partially because of their low tolerance to spatial misalignment. To this end, we propose the reference-oriented deformable convolution, which can improve the performance of an IQA network on GAN-based distortion by adaptively considering this misalignment.
We further propose a patch-level attention module to enhance the interaction among different patch regions, which are processed independently in previous patch-based methods. %
The modified residual block is also proposed by applying modifications to the classic residual block to construct a patch-region-based baseline called WResNet. Equipping this baseline with the two proposed modules, we further propose Region-Adaptive Deformable Network (RADN). The experiment results on the NTIRE 2021 Perceptual Image Quality Assessment Challenge dataset show the superior performance of RADN, and the ensemble approach won fourth place in the final testing phase of the challenge.

\end{abstract}
\vspace{-0.5cm}

\begin{figure}[t]
\begin{center}
  \includegraphics[width=1.0\linewidth]{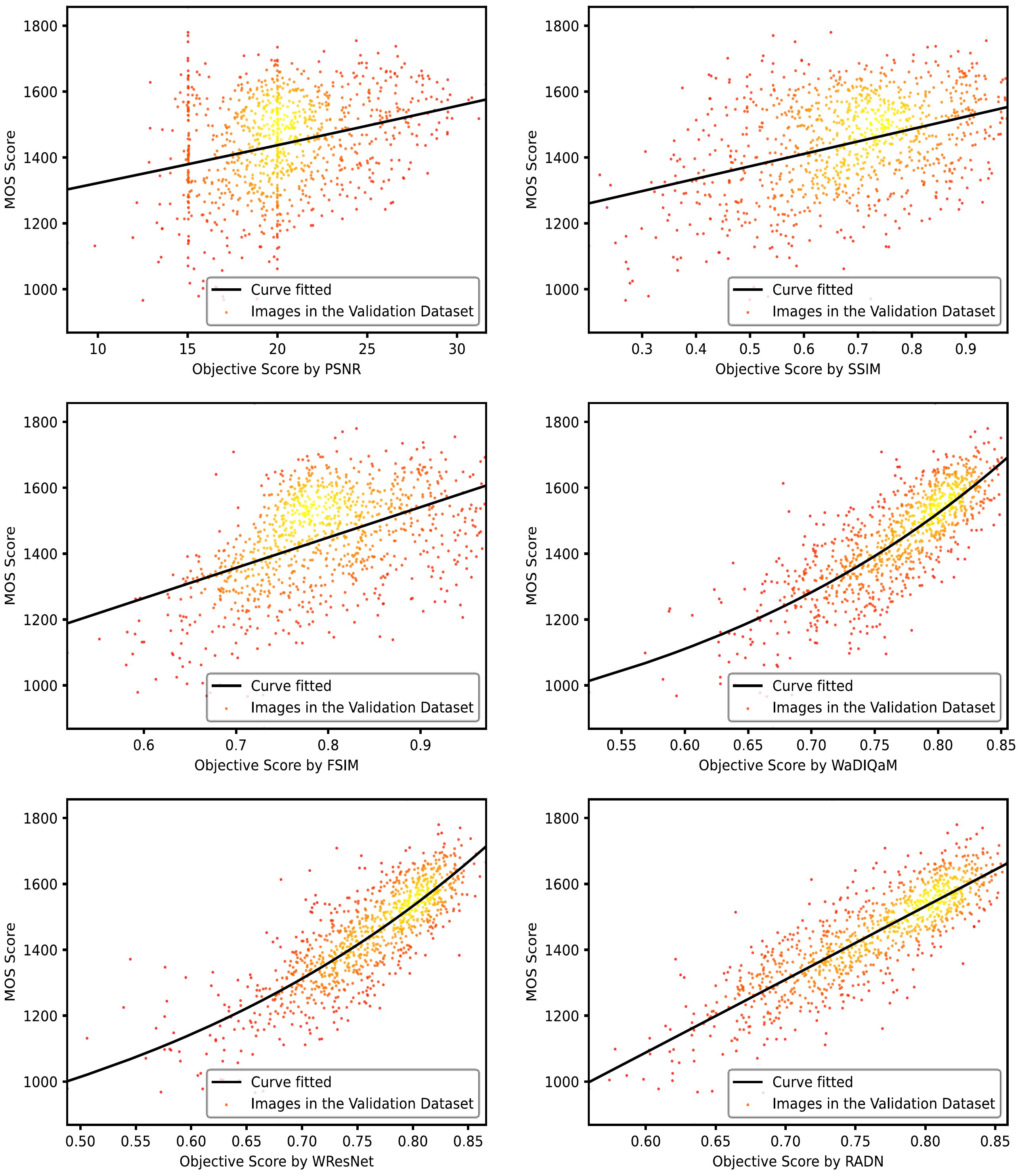}
\end{center}
\caption{The scatter plots of various IQA models on the NTIRE 2021 Perceptual Image Quality Assessment Challenge validation dataset, which show the relationship between the predicting scores and the MOS labels. } %
\label{fig:ScatterPlot}
\end{figure}

\vspace{-0.2cm}
\section{Introduction}
\label{sec:intro}
Image quality assessment tasks have gained increasing research attention for decades, and its goal is to assess image perceptual quality like humans. 
In the past decades, researchers have proposed some IQA algorithms based on deep learning~\cite{bosse2017deep,liu2017rankiqa,gu2020image,xia2020domain}. Although these IQA methods can maintain consistency with human subjective evaluation to some extent, they still show limitations in evaluating the results of image restoration and image super-resolution. 
As introduced in~\cite{gu2020image}, some GAN-based image restoration (IR) algorithms usually produce fake textures and other details. However, the existing algorithms cannot distinguish GAN-generated image textures from noises and natural details, which deteriorates the performance of existing IQA algorithms.
To deal with GAN-based distortion, Gu~\etal~\cite{gu2020image} proposed a novel IQA benchmark characterized by including a proportioned GAN-based distortion dataset, and most previously proposed IQA methods have shown unsatisfying performance on the dataset (see Fig.~\ref{fig:ScatterPlot}). They also propose a Space Warping Difference (SWD) layer to compare the features on a small range around the corresponding position. The operation is robust to spatial shifts. We observe that although the method has a specific effect, it cannot be used in all scenarios because the range of the field defined in~\cite{gu2020image} is a hyper-parameter that varies in different distortion scenarios. 
Therefore, it is limited to specific circumstances and is not general and flexible enough. 

Considering the drawbacks above, we propose the Region-Adaptive Deformable Network (RADN). The proposed method consists of three components: modified residual block, patch-level attention block, and reference-oriented deformable convolution.
We first revisit the classic residual blocks from IQA tasks and propose the modified residual block. 
Some modifications of the classical residual block are made to adapt the characteristics of IQA tasks, including removing the Batch Normalization (BN) layers, employing only 3 $\times$ 3 convolutional layers, and adjusting the numbers of the convolutional layers.
We then build our baseline WResNet using the modified residual blocks. Experiments in Sec.~\ref{section:as} show that the WResNet (without the other two modules) has already outperformed WaDIQaM~\cite{bosse2017deep} in both metric performance and convergence.

For adaptation to images of significant differences, we propose a novel module dubbed reference-oriented deformable convolution~\cite{dai2017deformable} which can select the region of interest adaptively according to the shape of the object.
Furthermore, it changes adaptively with the size of the target and selects critical information around it. 
We believe that the offset predicted in the reference image can capture the object's actual shape, similar to the result observed by humans, which cannot be affected by the GAN-based distortion. 
Applying such deformable convolution to the distorted images can make the reference images interact with the distorted images and be robust to the GAN-based distortion.

To boost the information interactions among the patch regions, we further propose a patch-level attention mechanism.
Despite good performance on synthetically distorted images of the patch-based algorithms, it may destroy both the high-level semantic information of the image and the relationship among patches. 
In other words, dividing a complete image into patches can affect the performance of the IQA methods. 
To alleviate this problem, we propose patch-level attention. 
The critical assumption is that the characteristics of an image patch depend not only on itself but also on other image patches. 
Therefore, we introduce patch-level attention after feature extraction of reference and distorted images to capture dependencies by computing interactions between two patch regions, respectively. 
This operation will strengthen the information interaction among different patches to obtain more accurate feature expressions for the two images. 
This plug-and-play module can be incorporated into any patch-based IQA method to improve performance.

Experiments on the newly-proposed PIPAL dataset and cross-dataset evaluation on TID2013 and LIVE show the competitive performance of our method on the above datasets. Our method ranked fourth in the NTIRE 2021 Perceptual Image Quality Assessment Challenge (NTIRE 2021 IQA Challenge)~\cite{gu2021ntire}.

\begin{figure*}[t]
\begin{center}
   \includegraphics[width=1.0\linewidth]{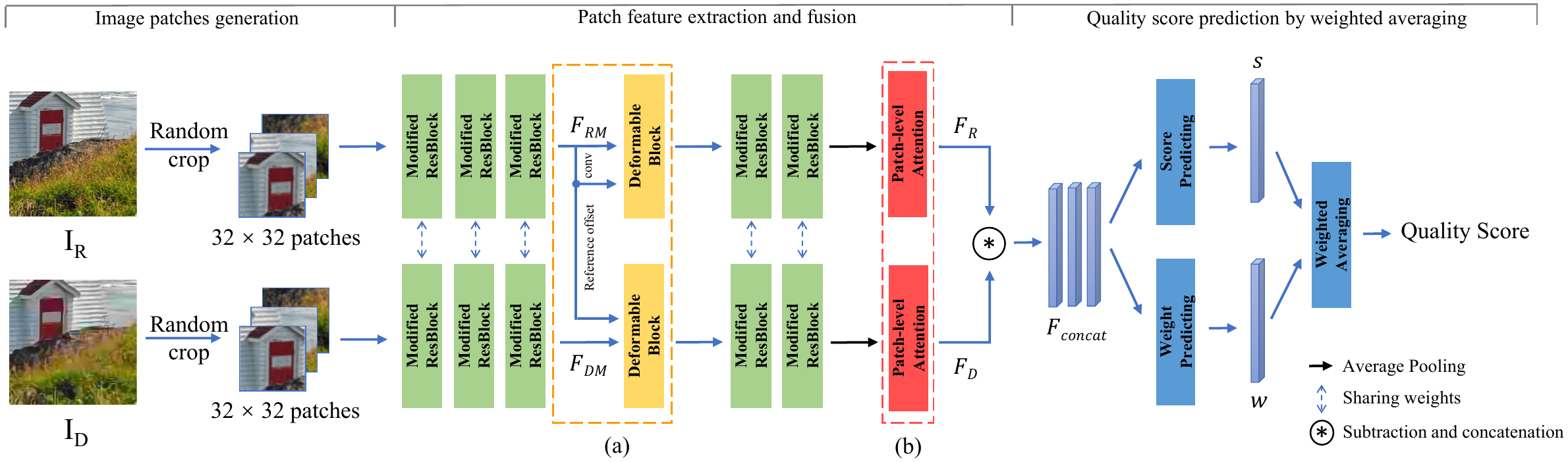}
\end{center}
    \caption{The architecture of the proposed approach - Region Adaptive Deformable Network (RADN). A distorted image $I_{D}$ and its corresponding reference image $I_{R}$ are randomly cropped into 32$\times$32 sized patches. Then the patches are inputted into the feature extracting module, and the final quality scores are obtained by weighted averaging.
    The yellow dashed box (a) and the red one (b) indicate our reference-oriented deformable convolution and patch-level attention module, respectively, which can be found in Sec.~\ref{subsec:deformableconv} and Sec.~\ref{subsec:plattention}. Our baseline model WResNet can be obtained by removing the two proposed modules (a) and (b). The details of our modified residual block can be found in Sec.~\ref{subsec:resblock}.
    }

\label{fig:RADN}
\end{figure*}

\section{Related Work}
\label{sec:related_work}
\noindent\textbf{Image Quality Assessment.}
The IQA algorithms are used to evaluate the quality of images that may be degraded during transmission, compression, and algorithm processing. Researchers have worked hard to develop general quality assessment algorithms close to human subjective evaluation in the past decades.  According to different scenarios, IQA algorithms can be divided into full-reference (FR-IQA) and no-reference methods (NR-IQA). FR-IQA methods commonly include SSIM \cite{SSIM}, MS-SSIM \cite{MS-SSIM} and PSNR, \etc. Inspired by them, FSIM \cite{FSIM}, SR-SIM \cite{SRSIM}, and GMSD \cite{xue2013gradient} are proposed. These hand-crafted methods assess the image quality by comparing the feature difference between the distorted image and the reference image. 
Recently, the deep learning-based FR-IQA methods~\cite{PIEAPP, bosse2017deep} get superior prediction performance over hand-crafted methods. Apart from FR-IQA methods, NR-IQA methods are developed to assess image quality without the reference image. Liu~\emph{et al.}~\cite{liu2017rankiqa} used comparative learning to assess image quality. In~\cite{zhu2020metaiqa}, meta-learning is introduced in IQA to learn meta-knowledge shared with humans.
In some recent works, IQA methods are applied to measure image restoration algorithms~\cite{PI}. Researchers hope to improve the performance of image restoration algorithms by developing better IQA methods. Gu~\emph{et al.}~\cite{gu2020image} proposed a new dataset named PIPAL. They test several existing algorithms to demonstrate that these algorithms' low tolerance toward spatial misalignment may be a key reason for their dropping performance. Unlike these approaches, we adopted patch-level attention and reference-oriented deformable convolution in our model to handle the GAN-based distorted images in PIPAL.

\noindent\textbf{Deformable Convolution.}
Dai~\emph{et al.}~\cite{dai2017deformable} first proposed deformable convolution and proved that it is effective for sophisticated vision tasks such as object detection~\cite{zhu2020deformable} and semantic segmentation~\cite{dai2017deformable}. Offsets learned in deformable convolution blocks can obtain the information in the light of the object's shape, improving the capability of the feature extraction. It also performs well in low-level vision tasks such as video super-resolution~\cite{tian2020tdan} and video deblurring~\cite{wang2019edvr}. However, it has not been introduced in the IQA task. Inspired by the methods mentioned above, we adopt reference-oriented deformable convolution for FR-IQA. 

\noindent\textbf{Attention Mechanism.}
Attention mechanisms have been widely used in various tasks~\cite{liu2017robust, liu2018non, vaswani2017attention, woo2018cbam}. For instance, in NR-IQA, Yang~\emph{et al.}~\cite{yang2019sgdnet} proposed an end-to-end saliency-guided architecture and applied spatial and channel attention in their model. Their method got a good performance in the NR-IQA task. Non-local operations \cite{wang2018non} compute the response at a position as a weighted sum of the features at all positions for capturing long-range dependencies. Motivated by these methods, we proposed patch-level attention to
capture dependencies between any two patches of one image to obtain more accurate feature maps from both reference and distorted images.

\section{Proposed Method}
\label{sec:method}

The structure of the proposed Region Adaptive Deformable Network (RADN) is shown in Fig.~\ref{fig:RADN}. 
For a pair of images, we first crop the reference image $I_R$ and the distorted image $I_D$ into patches with spatial size $32 \times 32$. 
During feature extraction, first, the modified residual blocks (green) are proposed and employed. Then with the intermediate feature maps of the reference and the distorted image (\ie, $F_{RM}$ and $F_{DM}$), we use the proposed reference-oriented deformable convolution (yellow) to select the region of interest and capture the object's actual shape adaptively. Next, the patch-level attention module (red) is employed to boost the interaction among the image patches. More details of the two proposed modules can refer to the Sec.~\ref{subsec:deformableconv} and Sec.~\ref{subsec:plattention}.
The final feature maps (\ie, $F_D$ and $F_R$) and their difference ($F_{diff}=F_D-F_R$) are then concatenated in the channel dimension to serve as a new feature, \ie $F_{concat}=concat(F_{diff}, F_D, F_R)$. Finally as in Eq.\ref{eq:weightedAveraging}:

\begin{equation}
\label{eq:weightedAveraging}
    \hat{q} = \dfrac{\sum_{\substack{0<i<N_{patch}}}w_{i} \times s_{i} }    
              {\sum_{\substack{0<i<N_{patch}}} w_{i}}
\end{equation}

The combined feature $F_{concat}$ will be sent to the fully-connected layers to predict the weight $w_{i}$ and the score $s_{i}$ of per patch and get the final quality score $\hat{q}$ by weighted averaging and $N_{patch}$ in the equation means the amount of the patches for one image. 

Besides the aforementioned modules, we also propose a contrastive pretraining strategy to further improve the model's ability to distinguish the image quality rather than direct regression of the quality score.

\subsection{Modified Residual Block}
\label{subsec:resblock}

\begin{figure}[t]
\begin{center}
\includegraphics[width=0.85\linewidth]{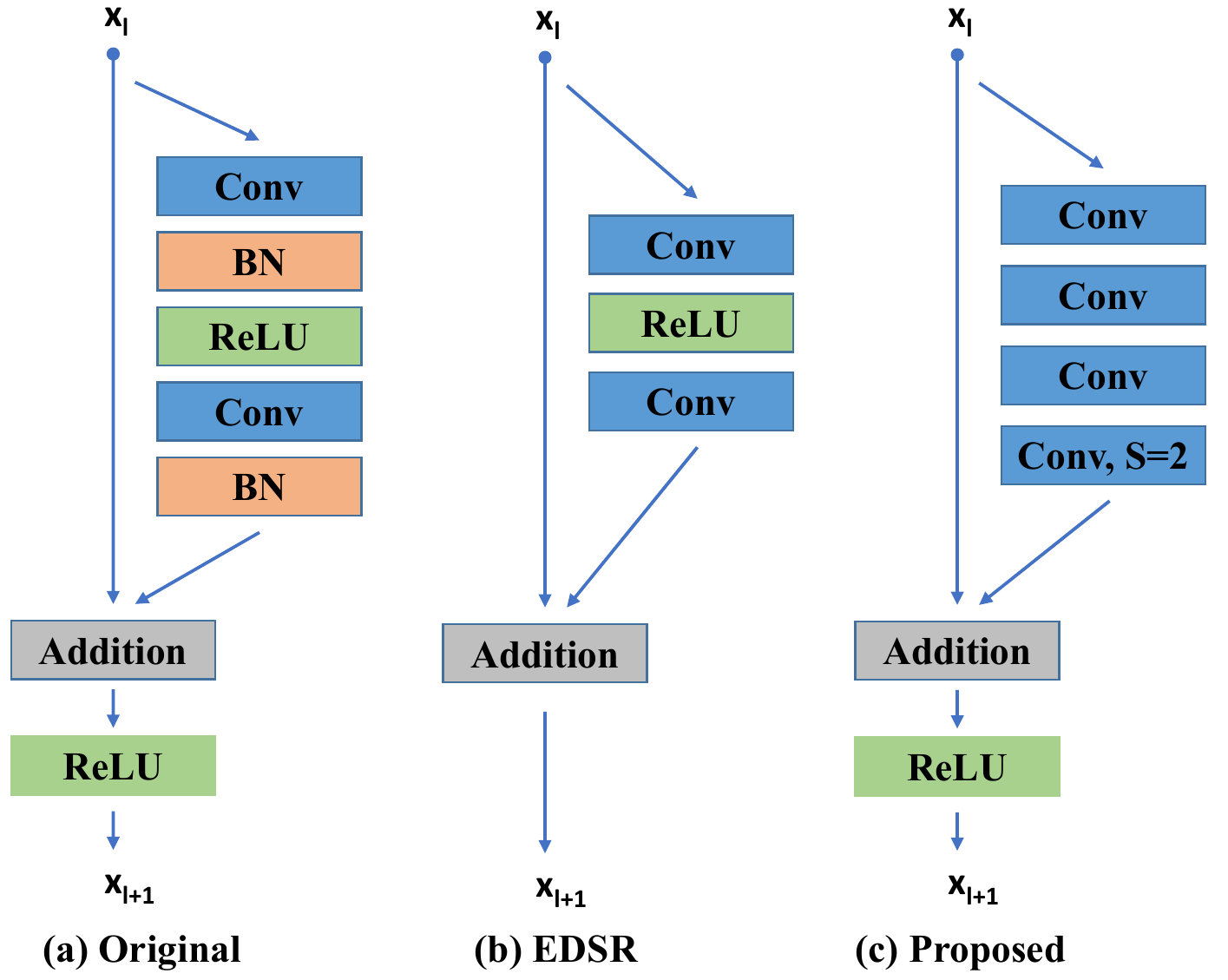}
\end{center}
\caption{The architectures of various residual blocks.
Note that we use 2-stride convolution layers of the residual blocks to perform down-sampling operations.}
\label{fig:ResBlock}
\end{figure}

Considering the characteristics of IQA tasks and the PIPAL dataset~\cite{gu2020image}, we improve the performance of the classic residual block with a more reasonable structure (shown in Fig.~\ref{fig:ResBlock}).
The low-quality images vary greatly in content and distortion types. 
As batch normalization operations will result in over-smoothness of special features in different samples~\cite{Lim2017enhanced,wang2018esrgan} which makes the model performance degrade greatly, we remove all the batch- normalization layers in the original residual block. 

Furthermore, we adopt the $3 \times 3$ convolution without any $7 \times 7$ employed in the original ResNet. The 3 $\times$ 3 convolution has been proven to be more hardware-friendly~\cite{ding2021repvgg} and more effective than the $7 \times 7$ convolution layer followed by a pooling layer in the earliest feature extraction stage. 
We only add a shortcut every four convolutional layers, which can generate more complex representations and show better performance in our experiment than the two convolution layers adopted in the original residual block.

\begin{figure}[t]
\begin{center}
   \includegraphics[width=0.9\linewidth]{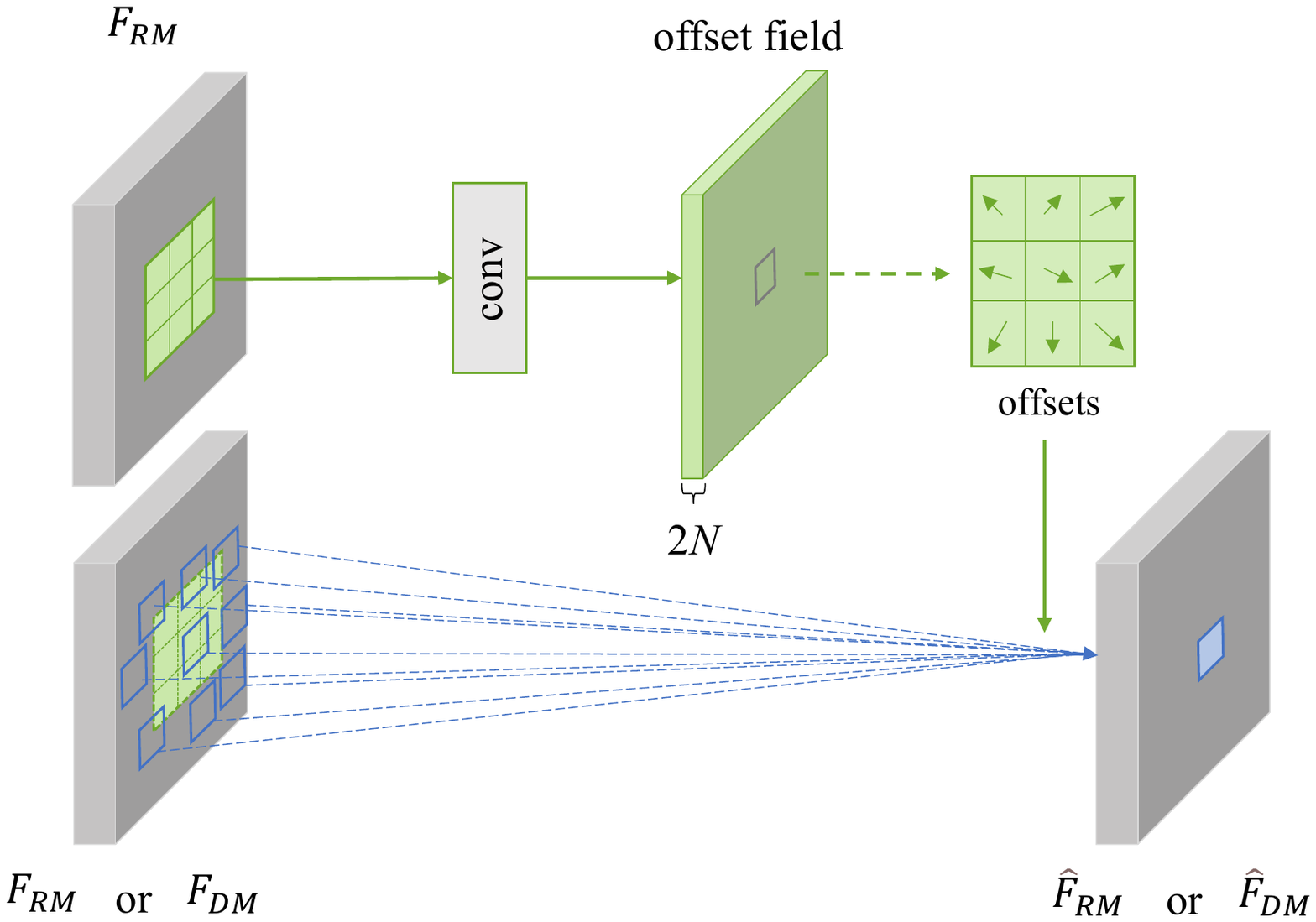}
\end{center}
   \caption{Reference-oriented deformable convolution. $F_{RM}$ and $F_{DM}$ indicate the feature map of the reference image and the distorted image respectively. The number of offsets is defined as: $N$= $|\mathbf{p}_k|$.}
\label{fig:Deformable Block}
\end{figure}

\subsection{WResNet}
\label{subsec:wresnet}
We use the modified residual blocks to build our baseline method named WResNet (W stands for weighted averaging). The architecture of our baseline is shown in Fig.~\ref{fig:RADN}, which can be obtained by removing the two proposed modules (a) and (b). For WResNet, we apply the $l_{2}$ loss function to regress the quality scores during training.

\subsection{Reference-Oriented Deformable Convolution}
\label{subsec:deformableconv}
Considering that humans are less sensitive to the error and misalignment of the edges in distorted images generated by GAN, Gu~\emph{et al.}~\cite{gu2020image} proposed SWD to deal with the GAN-based distortion. In fact, the fixed design limits its adaptation to different GAN-based distortion. Therefore, we adopt the deformable convolution module in our residual blocks to adapt to the mismatched regions of GAN-based distortion between reference and distorted images. To make better use of the reference information, different from the original deformable convolution, we introduce reference-oriented one shown in Fig.~\ref{fig:Deformable Block}, where $F_{DM}$ and $F_{RM}$ are the intermediate distorted and reference feature maps (see Fig.~\ref{fig:RADN}). For conventional 2D convolution with a kernel of $N$ sampling locations, it first samples the values from the regular offsets $\mathbf{p}_k, k \in {1, 2, \cdots, N}$, then sums the sampled values weighted by $W_k$ corresponding to the $k$th location. For example, a 3×3 kernel has 9 regular sampling locations which are defined as $\mathbf{p}_k{\in}\{(-1,-1),(-1,0),\cdots,(1,1)\}$. In terms of our reference-oriented deformable convolution, we first generate the offsets $\Delta \mathbf{p}_k$ according to $k$th sampling location from reference feature maps:
\vspace{-0.1cm}
\begin{equation}
    \Delta \mathbf{p}_k=f(\mathbf{F}_{RM})
\vspace{-0.1cm}
\end{equation}
where $f$ means a $3 \times 3$ convolution, the output channel number of which is $2N$. 
Then the learned offsets $\Delta \mathbf{p}_k$ are used for sampling the values in both reference and distorted feature maps, rather than the regular sampling with $\mathbf{p}_k$.
For each location $\mathbf{p}_0$ on the output reference and distorted feature maps $\hat{F}_{RM}$ and $\hat{F}_{DM}$, the value in it is aggregated by the following process:
\vspace{-0.2cm}
\begin{align}\label{equ:rdcn}
\hat{\textbf{F}}_{RM}(\mathbf{p}_0) &= \sum_{k=1}^{N} w^r_k\cdot \mathbf{F}_{RM}(\mathbf{p}_0+\mathbf{p}_k+\Delta \mathbf{p}_k), \nonumber \\
\hat{\mathbf{F}}_{DM}(\mathbf{p}_0) &= \sum_{k=1}^{N} w^d_k\cdot \mathbf{F}_{DM}(\mathbf{p}_0+\mathbf{p}_k+\Delta \mathbf{p}_k),
\vspace{-0.1cm}
\end{align}
where $w^r$ and $w^d$ are the convolution weights for reference and distortion feature maps. 
Owing to the deformable sampling locations $\mathbf{p}_0+\mathbf{p}_k+\Delta \mathbf{p}_k$ are fractional, bilinear interpolation is applied~\cite{dai2017deformable} to sample the values. 

With the application of this reference-oriented deformable convolution in both reference and distorted branches (shown in Fig.~\ref{fig:RADN}), our model can deal with the GAN-based distortion better and learn the spatial shift-invariant features from the paired images adaptively. 
We demonstrate the effectiveness of reference-oriented deformable convolution in Sec.~\ref{section:as}. 

\begin{figure}[t]
\begin{center}
   \includegraphics[width=0.85\linewidth]{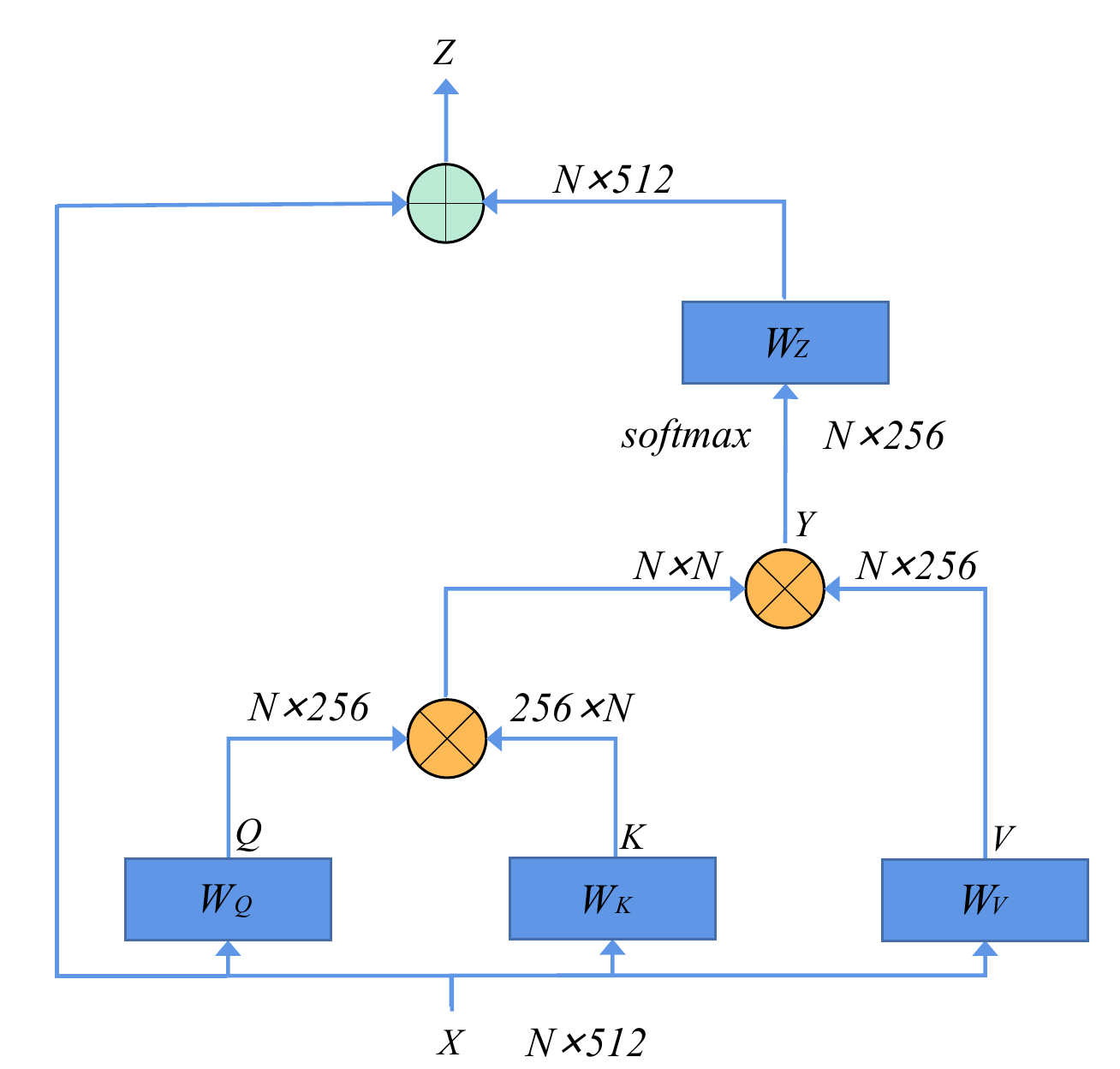}
\end{center}
   \caption{The patch-level attention block. The feature maps are shown as the shape of their tensors, and $N$ is the number of patches.}
\label{fig:Patch-level Attention Block}
\end{figure}

\subsection{Patch-Level Attention}
\label{subsec:plattention}
The previous patch-level-based IQA methods predict the quality of each patch individually, which leads to a specific performance drop due to the lack of interaction among the patch regions. 
We believe the quality of a patch not only depends on its own feature but also affected by other patches in the same image. Recently, the self-attention mechanism exhibits excellent relation modeling in computer vision tasks from low-level to high-level tasks. However, how to introduce it into the FR-IQA task is a challenge to be explored. To this end, we introduce a self-attention module to handle the feature maps of reference and distorted images to boost the interaction between any two patches in one image, as shown in the red dashed box in Fig.~\ref{fig:RADN}. 
Different from the non-local block proposed in \cite{wang2018non}, we compute the response at a patch as a weighted sum of the features at all patches in one image, namely patch-level attention.

Since dot-product attention is much faster and more space-efficient in practice, we use it scaled by $\frac{1}{\sqrt{d_k}}$ in our patch-level attention block.
Attention block generates corresponding \textbf{Q}ueries, \textbf{K}eys and \textbf{V}alues of dimension ${d}_k$ by performing linear projection on the patch-level feature vectors $X$ with $W_{Q}$, $W_{K}$, $W_{V}$:
\begin{equation}\label{eq:matrix compute}
    Q = W_Q X, \quad  K= W_K X, \quad V=W_V X 
\end{equation} 
where $X$ is reshaped from $N \times 512 \times 1 \times 1$ to $N \times 512$ ($N$ is the number of patches) before it is input into the patch-level attention block. Attention operations are defined as following:
\begin{equation}\label{eq:attention}
   Y = \mathrm{softmax}(\frac{QK^T}{\sqrt{d_k}})V
\end{equation}
We involve the attention operation described in Eq.~\ref{eq:attention} into a patch-level attention block that can be integrated into our model. It is defined as:
\begin{equation}\label{eq:block}
    Z = W_Z Y + X
\end{equation}
where $Y$ is given in Eq.~\ref{eq:attention}, $W_Z$ is a weight matrix to be learned and $Z$ is computed by a residual connection. An example of patch-level attention block is shown in Fig.~\ref{fig:Patch-level Attention Block}. 

Through the processing of the patch-level attention block, the patches in the same image can be better interacted with each other and have more accurate feature representation.

\begin{figure}[t]
\begin{center}
    \includegraphics[width=1.0\linewidth]{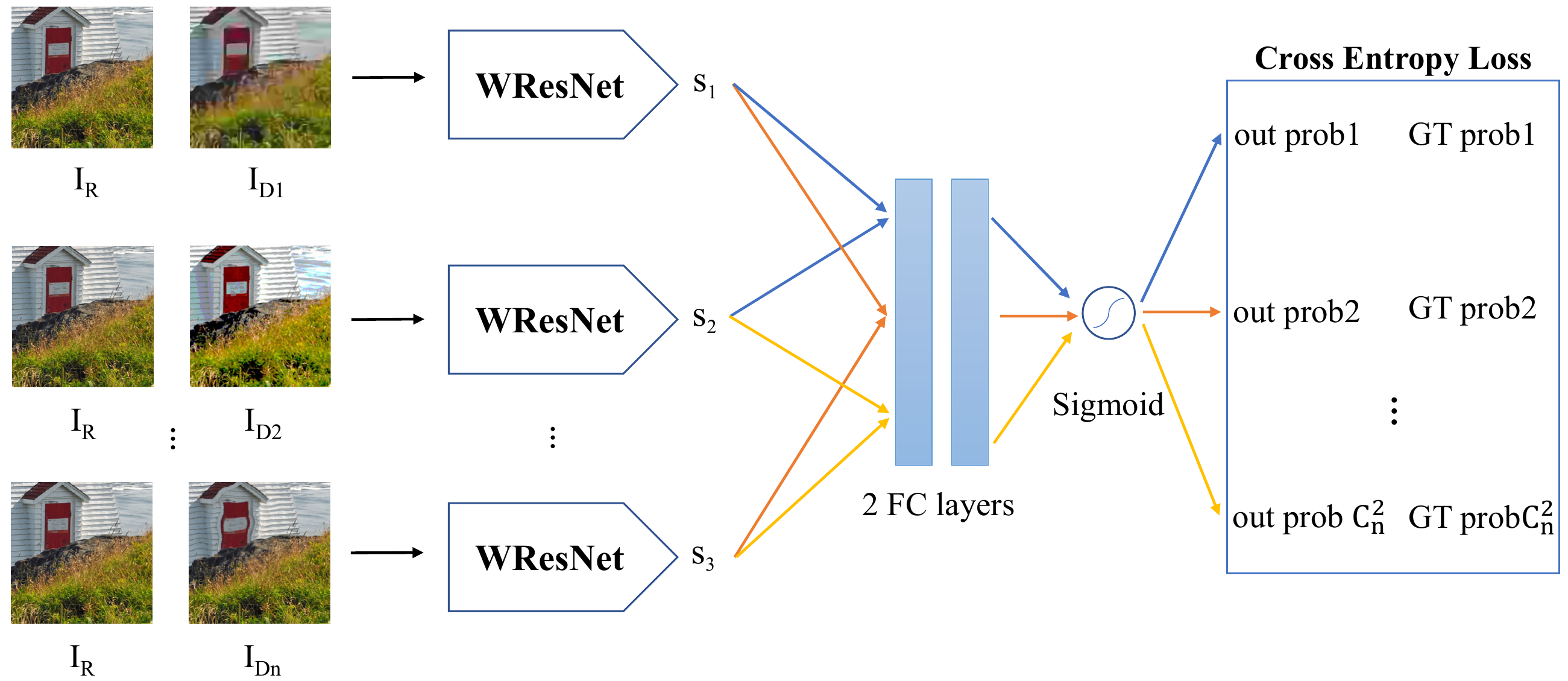}
\end{center}
   \caption{The architecture of our efficient contrastive model. Arrows with the same color stand for a contrast pair and its result. Every two scores are compared as a contrast pair so if there are $n$ pairs of reference and distorted images, there will be $C_{n}^2$ contrast pairs and preference probabilities.}
\label{fig:Compare}
\end{figure}
\subsection{Contrastive Pretraining Strategy}
\label{subsec:pretrain}
Contrastive training is an acceptable way to take advantage of the labels from the side, given that the MOS labels are obtained by manually comparing the image pairs. 
Most current IQA models tend to use $l_{1}$ and $l_{2}$ to regress the quality score, which merely concentrates on the accuracy of the values and ignores the ranking relationships between the samples. 
The contrastive models~\cite{gu2020image,liu2017rankiqa} can alleviate this problem by comparing the samples.
Based on SWDN~\cite{gu2020image}, we propose a contrastive pretraining strategy to make the model learn how to distinguish the image quality rather than directly regress the quality score. 
The difference is that our strategy is to pre-train the model by comparing the labels to get the preference probabilities and then make use of the MOS labels directly by $l_{2}$ regression.
Siamese Network is an indispensable part of contrastive learning, and here we use our WResNet proposed in Sec.~\ref{subsec:wresnet}.
As in Fig.~\ref{fig:Compare}, given a set of $n$ quality scores $\{s_{1}, s_{2}, \cdots, s_{n}\}$ obtained by the Siamese Network, we make a comparison between every two samples $(s_{i},s_{j})$ using the fully-connected layers $f(s_{i},s_{j})$ to get the preference probabilities. 
Then we apply the cross-entropy loss to regress them with the ground truth preferring probability $p_{ij}$. 
The final cross-entropy loss is shown as follows:

\begin{equation}
  \begin{split}
    L(s_{i},s_{j},p_{ij})=\sum_{\substack{0<i<n\\i<j<n}}   
    -p_{ij} \times f(s_{i},s_{j}) \\ - (1-p_{ij}) \times (1-f(s_{i},s_{j}))
  \end{split}
\end{equation}
We implement the efficient back-propagation in Siamese networks proposed by RankIQA~\cite{liu2017rankiqa} to remarkably improve the training efficiency \ie, from 5-7 hours per epoch to 20 minutes per epoch, and improve the performance of the pretrained model. 
We validate the effectiveness of our contrastive pretraining strategy in Sec.~\ref{section:as}. %

\section{Experiments}

\begin{table}[t]
\centering
\caption{Performance of different methods on the NTIRE 2021 IQA Challenge validation and testing datasets. The results of our ensemble model is bolded.
}
\label{tab:val_table}
\resizebox{0.9\linewidth}{!}{
\begin{tabular}{ccccc}
\toprule

\multicolumn{1}{c|}{\multirow{2}{*}{Method}} & \multicolumn{2}{c}{Validation} & \multicolumn{2}{c}{Test} \\ \cline{2-5} 
\multicolumn{1}{c|}{}                        & SROCC          & PLCC          & SROCC       & PLCC       \\ \hline
PSNR         & 0.2548  & 0.2917 & 0.2493 & 0.2769\\ %
NQM \cite{NQM}         & 0.3458  & 0.4164 & 0.3644 & 0.3954\\
UQI \cite{UQI}          & 0.4859  & 0.5476 & 0.4195 & 0.4500\\
SSIM \cite{SSIM}         & 0.3400  & 0.3984 & 0.3614 & 0.3936\\
MS-SSIM \cite{MS-SSIM}      & 0.4864  & 0.5633 & 0.4618 & 0.5007\\
IFC \cite{IFC}           & 0.5936  & 0.6767 & 0.4851 & 0.5549\\
VIF \cite{VIF}          & 0.4335  & 0.5236 & 0.3970 & 0.4795\\
VSNR \cite{VSNR}         & 0.3213  & 0.3750 & 0.3682 & 0.4107\\
RFSIM \cite{RFSIM}        & 0.2656  & 0.3045 & 0.3037  & 0.3284\\
GSM \cite{GSM}         & 0.4181  & 0.4688 & 0.4094 & 0.4646\\
SRSIM \cite{SRSIM}       & 0.5658  & 0.6541 & 0.5728 & 0.6360\\
FSIM \cite{FSIM}        & 0.4672  & 0.5606 & 0.5038 & 0.5709\\
FSIMc \cite{FSIM}       & 0.4679  & 0.5587 & 0.5057 & 0.5727\\
VSI \cite{VSI}         & 0.4501  & 0.5162 & 0.4584 & 0.5169\\
MAD \cite{MAD}         & 0.6078  & 0.6263 & 0.5434 & 0.5804\\
NIQE \cite{NIQE}        & 0.0644  & 0.1018 & 0.0341 & 0.1317\\
MA \cite{MA}          & 0.2006  & 0.2034 & 0.1405 & 0.1469\\
PI \cite{PI}          & 0.1690  & 0.1662 & 0.1036 & 0.1454\\
\hline
LPIPS-Alex \cite{LPIPS} & 0.6276  & 0.6463 & 0.5658  & 0.5711\\
LPIPS-VGG \cite{LPIPS}  & 0.5915  & 0.6471 & 0.5947 & 0.6331\\
PieAPP \cite{PIEAPP}      & 0.7063 & 0.6972 & 0.6074 & 0.5974\\
WaDIQaM \cite{bosse2017deep}     & 0.6779 & 0.6543 & 0.5533 & 0.5480\\
DISTS \cite{DISTS}       & 0.6743  & 0.6860  & 0.6548 & 0.6873\\
SWD \cite{gu2020image}         & 0.6611  & 0.6680 & 0.6243 & 0.6342\\
WResNet-Classic              & 0.4881  & 0.4868 & 0.4891 & 0.5056\\
WResNet-EDSR              & 0.6920  & 0.6856 & 0.6789 & 0.6877\\
WResNet              & 0.8137  & 0.8177 & 0.7501 & 0.7542\\
\textbf{Ours}        & \textbf{0.8655}  & \textbf{0.8666}  & \textbf{0.7770}  & \textbf{0.7709}\\

\bottomrule
\end{tabular}
}
\end{table}

\subsection{Datasets}
\label{subsec:database}
PIPAL~\cite{jinjin2020pipal} is a recently proposed IQA dataset, which contains images processed by image restoration and enhancement methods (particularly the deep learning-based methods) besides the traditional distorting methods.
The dataset contains 250 reference images, 29k distorted images of 40 distortion types, and 1.13m human judgment quality scores. 
We also conduct cross-dataset evaluation on TID2013~\cite{ponomarenko2015image} and LIVE~\cite{sheikh2006statistical}.

\subsection{Implementation Details}

\noindent\textbf{Training Details.}
We train the proposed model with the patch-based training strategy in WaDIQaM~\cite{bosse2017deep}, which has been proved to be able to augment the data effectively and improve the performance. 
For training, we randomly sample 32 patches from per distorted image and its corresponding reference image rather than the whole image. These sampled patches could be overlapped. We set the mini-batch size as 2 for the consideration of the remarkable difference between distorted images. We found in experiments that the model can better learn and adapt to such differences with small batch sizes. We use ADAM optimizer with the parameters $\beta_1= 0.9$, $\beta_2= 0.999$.
For our models, the learning rate is initialized as $10^{-4}$ and will decay to 0.8 of itself at every 100 epochs.
For testing, each pair of images are cropped into a certain number $M$ of 32 $\times$ 32 \textit{non-overlapping} patches. These patches are then fed into the network to predict the weight $w_i$ and the score of each patch $s_i$. The final quality score $s$ for the distorted image is calculated by $s=\sum_{i=1}^{M}w_is_i$.

\begin{figure}[t]
\begin{center}
   \includegraphics[width=1.0\linewidth]{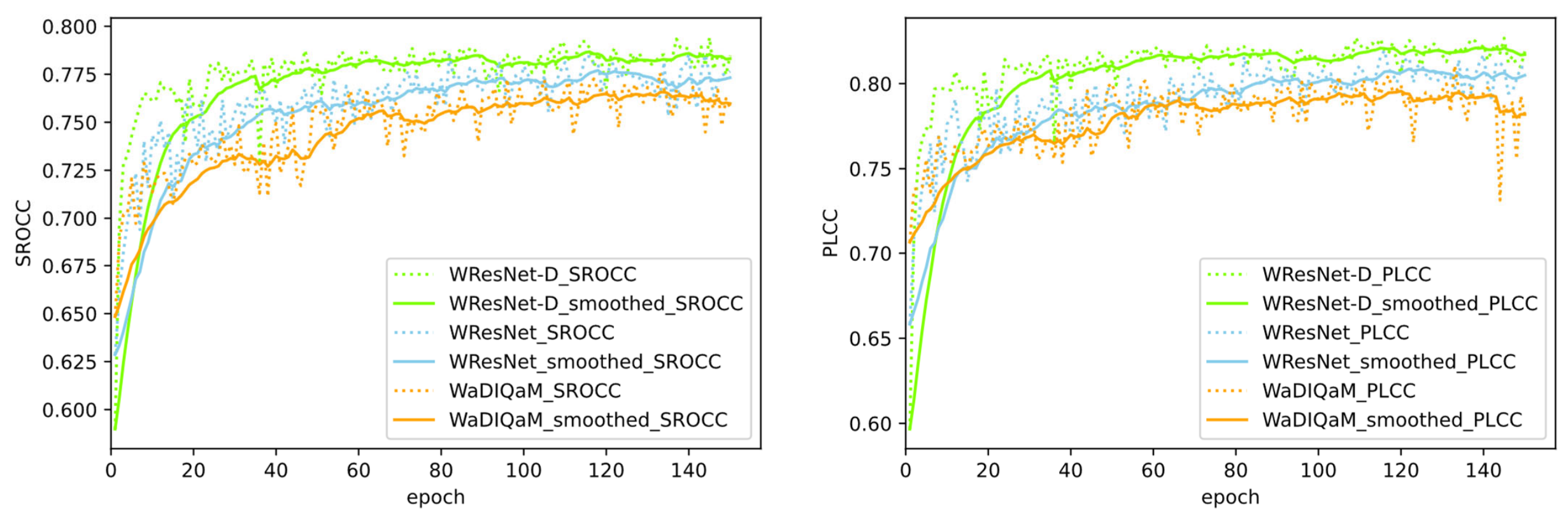}
\end{center}
   \caption{SROCC and PLCC performance of various models on our validation dataset. The solid curves are the smoothed ones with inertial filtering while the dotted curves are the original ones without smoothing. WResNet-D indicates WResNet with reference-oriented deformable convolutions.}
\label{fig:SROCC}
\end{figure}

\begin{table}[t]
\caption{Cross-dataset evaluation on TID2013 and LIVE.}
\label{tab:cross_validation}
\resizebox{0.95\linewidth}{!}{

\begin{tabular}{ccccc}
\toprule

\multicolumn{1}{c|}{\multirow{2}{*}{Method}} & \multicolumn{2}{c}{TID2013} & \multicolumn{2}{c}{LIVE} \\ \cline{2-5} 
\multicolumn{1}{c|}{}                        & SROCC         & PLCC        & SROCC       & PLCC       \\ \hline
PSNR                                         & 0.687        & 0.677           & 0.873         & 0.865
    \\
WaDIQaM                                      & 0.698            & 0.741            & 0.883            & 0.837           \\
RADN                                         & 0.747              & 0.796             & 0.905             & 0.878            \\ 
\bottomrule
\end{tabular}
}
\end{table}

\begin{figure*}[]
\begin{center}
   \includegraphics[width=1.0\linewidth]{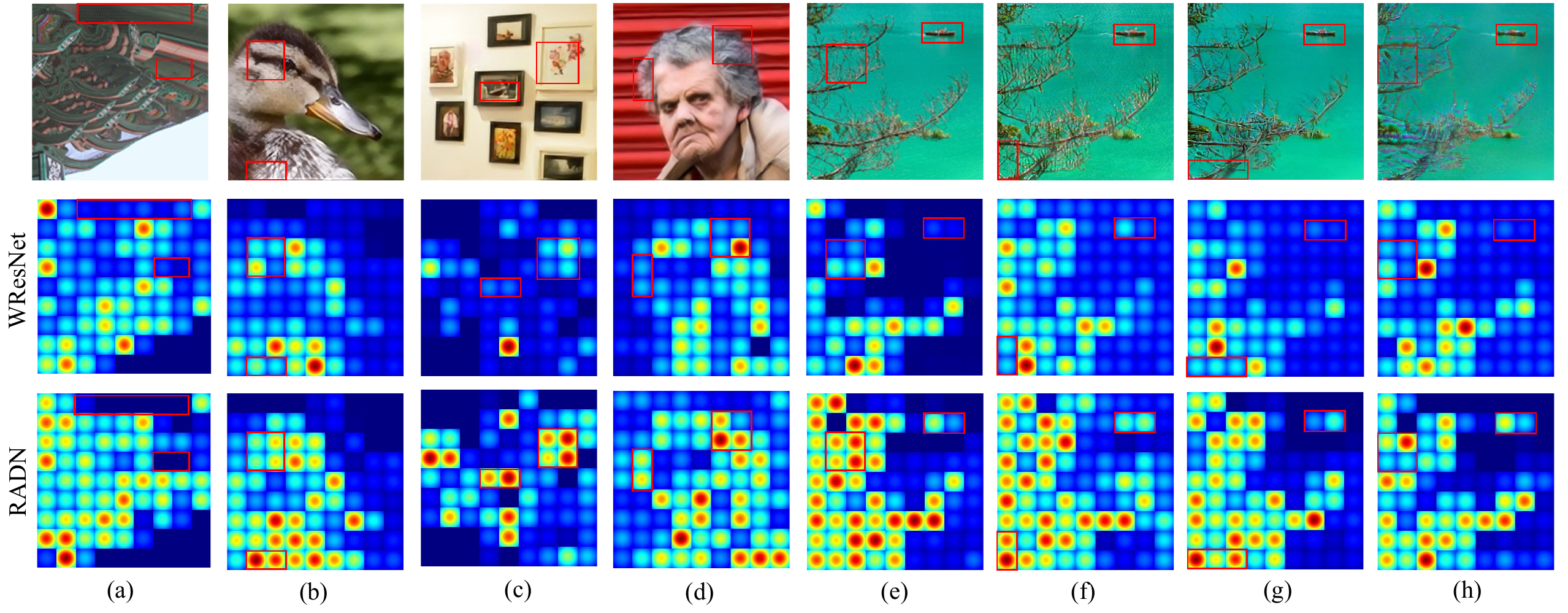}
\end{center}
   \caption{The attention maps of the images obtained by visualizing the weights of the patches. 
   The distortion types of (a)-(e) are traditional, and the rest are GAN-based. 
   The comparison between our proposed RADN and WResNet shows the effectiveness of our proposed modules, which contributes to focusing on the noteworthy regions for IQA tasks, especially for the GAN-based distortions.}
\label{fig:WeightAttention}
\end{figure*}

\noindent\textbf{Data Arrangement for Contrastive Pretraining.}
As mentioned in Sec.~\ref{subsec:database}, the PIPAL dataset contains 7 distortion categories and 40 distortion subtypes.
Each reference image corresponds to 116 distorted images of different sub-types from the seven distortion categories, but the specific subtype is unknown.
Considering the gap among various images, we collect distorted images corresponding to the same reference and from the same category for contrastive pretraining. 
We call these image collections `contrast groups'.
To implement the efficient Siamese Network, we arrange the images from the same contrast group into one batch to avoid the duplicate score computing process. Thus, the training time would be essentially saved.

\subsection{Evaluation Criteria}
To evaluate the performance, we use Spearman's rank order correlation coefficient (SROCC) and Pearson's linear correlation coefficient (PLCC), following the prior works~\cite{gu2020image,bosse2017deep,liu2017rankiqa}.
For $N$ testing images, the PLCC is defined as follows:
\begin{equation}
\label{eq11}
  \textrm{PLCC} = \frac{\sum_{i=1}^{N}(s_i - \mu_{s_i})(\hat{s}_i - \mu_{\hat{s}_i})}{\sqrt{\sum_{i=1}^{N}(s_i - \mu_{s_i})^2}\sqrt{\sum_{i=1}^{N}(\hat{s}_i - \mu_{\hat{s}_i})^2}} 
\end{equation}
Where $s_i$ and $\hat{s}_i$ respectively indicate the ground-truth and predicted quality scores of $i$-th image, and $\mu_{s_i}$ and $\mu_{\hat{s}_i}$ indicate the mean of them. Let $d_i$ denote the difference between the ranks of $i$-th test image in ground-truth and predicted quality scores. The SROCC is defined as

\begin{equation}
\label{eq12}
  \textrm{SROCC} = 1 - \frac{6\sum_{i=1}^{N}d_i^2}{N(N^2-1)} 
\end{equation}
Both metrics, PLCC and SROCC, are in [-1, 1], and higher values indicate better performance.

\subsection{Comparison with the State-of-the-arts}
\label{subsec:comparison}
\noindent\textbf{Evaluation on PIPAL.}
We compare our models with the state-of-the-art FR-IQA methods on the NTIRE 2021 IQA challenge validation and testing datasets. 
The quantitative comparisons on both datasets are shown in Tab.~\ref{tab:val_table} and the 'ours' term indicates our ensemble approach. 
We divide the general-purpose IQA methods into traditional methods and deep learning-based methods. In general, deep learning-based methods achieved better performance than the traditional methods. As can be seen, our method is superior to WaDIQaM~\cite{bosse2017deep} which is also a patch-based approach and widely used in the assessment of synthetically distorted images. Our method especially achieves superior results over PieAPP~\cite{PIEAPP} which is considered the most effective approach in~\cite{jinjin2020pipal} on both datasets by a large margin. 

\noindent\textbf{Cross-dataset Evaluation on TID2013 and LIVE.}
To validate the generalization of our proposed RADN, we conduct the cross-dataset evaluation on TID2013 and LIVE. We trained RADN on the training set of PIPAL and test it on the full set of TID2013 and LIVE.
As shown in Tab.~\ref{tab:cross_validation}, RADN outperforms WaDIQaM~\cite{bosse2017deep} and PSNR with large margins on both datasets, which indicates the effectiveness of our proposed modules.

\subsection{NTIRE2021 IQA Challenge}
\label{sec:challenge}
Our methods are originally proposed for participating in the NTIRE 2021 Perceptual Image Quality Assessment Challenge\cite{gu2021ntire}, which aims to establish an algorithm to measure the visual quality of the images fairly and focuses on the PIPAL dataset. Our ensemble approach with the strategies mentioned above ranked 3rd place in the public validation phase (also called the development phase) and ranked 4th place in the final private test phase (as shown in Tab.~\ref{tab:val_table}).

\begin{table}[t]
\caption{Ablation study on the validation dataset of the NTIRE 2021 IQA Challenge. \textit{Contrastive} refers to our contrastive pretraining strategy. \textit{Deform} indicates our reference-oriented deformable module and \textit{PatchAttn} indicates our patch-level attention module.}
\label{tab:ablationStudy}
\centering
\resizebox{0.95\linewidth}{!}{
\begin{tabular}{ccc|cc}
\toprule
Contrastive & Deform & PatchAttn &  SROCC  &  PLCC    \\
\hline
            &        &         & 0.8137  & 0.8177  \\
    $\surd$ &        &         & 0.8244  & 0.8252  \\
    $\surd$ & $\surd$&         & 0.8329  & 0.8337  \\
    $\surd$ &        & $\surd$ & 0.8343  & 0.8345  \\
    $\surd$ & $\surd$& $\surd$ & \textbf{0.8438}  & \textbf{0.8435} \\

\bottomrule
\end{tabular}
}
\end{table}

\subsection{Ablation Study} 
\label{section:as}
To further investigate the effectiveness of our proposed components, we conduct ablation studies on the validation dataset of the NTIRE 2021 IQA Challenge. 
Both SROCCs and PLCCs are shown in Tab.~\ref{tab:ablationStudy}.

\noindent\textbf{Modified Residual Block.}
Fig.~\ref{fig:SROCC} depicts the performance of different models on the validation dataset during training. 
The dotted curves are the original ones without smoothing, while the solid curves are the smoothed ones obtained by inertial filtering.
The orange curves indicate WaDIQaM, and the blue ones indicate WResNet with our modified residual blocks.
We can easily conclude that WResNet outperforms WaDIQaM on SROCC and PLCC. During the training process, our WResNet ascend more steadily compared with the vibrated curve of WaDIQaM.
Also as in Tab.~\ref{tab:val_table}, WResNet-Classic and WResNet-EDSR indicate WResNet with classic and EDSR-like residual blocks respectively. For a fair comparison, all three models adopt 20 convolution layers, and the superior performance of WResNet demonstrates the effectiveness of our modified residual block. The architectures of various residual blocks can refer to Fig.~\ref{fig:ResBlock}.

\noindent\textbf{Contrastive Pretraining Strategy.}
The contrastive training strategy is adopted for pretraining our models. 
Tab.~\ref{tab:ablationStudy} shows that the proposed pretraining strategy can learn the contrastive knowledge priors, which can improve the model's performance.

\noindent\textbf{Reference-oriented Deformable Convolution.}
We only add reference-oriented deformable convolution to our baseline (\ie, {WResNet+Deform}), a major improvement on SROCC and PLCC are shown in Tab.~\ref{tab:ablationStudy}. The training process (indicate by the green curves) in Fig.~\ref{fig:SROCC} also shows superior performance compared with our baseline model, which indicates the proposed reference-oriented deformable convolution module can adapt to the GAN-based distortion scenarios in PIPAL. 

\noindent\textbf{Patch-Level Attention.}
The patch-level attention mechanism is proposed to enhance interactions among all patches. The results in Tab.~\ref{tab:ablationStudy} show such interactions are indispensable for the patch-based IQA algorithms. To be specific, the \emph{patch-level attention module} gains another 0.01 improvements on SROCC and PLCC, respectively. 

Combined with all the components we proposed, RADN significantly improves the evaluation performance, especially compared to our baseline.

\subsection{Visualization and Discussion}
To intuitively illustrate the effectiveness of our method, we visualize the weights of patches in some images, as shown in Fig.~\ref{fig:WeightAttention}. 
According to the weight from the lowest to the highest, the color of the patch is displayed as blue, green, yellow, and red. A higher weight means the model pays more attention to the patch region.
The distortion types of (a)-(e) are traditional, and the rest are GAN-based. %
The second and third-row visualization results come from the WResNet (without the deformable and patch-level attention modules) and the whole method RADN. 

As shown, both methods can adaptively give higher weights to noteworthy regions. 
These regions usually contain complex textures or salient subjects, which are essential in assessing the image quality because human tends to pay more attention to such regions.
The highlighted regions are indicated with red boxes.
Compared with WResNet, our whole model RADN pays less attention to the regions with less informative or the texture-less regions as shown in Fig.~\ref{fig:WeightAttention} (a).
As shown in Fig.~\ref{fig:WeightAttention} (b)-(e), for images of different content, RADN perceives the images better and observably stays under human's perception - for example, the contours of the duck's head, the paintings, and the elder's head are clearly outlined, which strongly shows the effectiveness of our proposed modules.

To further illustrate the effectiveness of our method on GAN-based distortion, we show the visualized results for different GAN-based distortion images according to the same reference in Fig.~\ref{fig:WeightAttention} (f)-(h).
Despite the diversity and severe GAN-based distortions, our RADN can capture the actual outline of the attracting targets like the boat and the branches of the tree because of the reference-oriented deformable convolution. Also, RADN distributes fewer attention weights in flat areas like the sea surface due to the effectiveness of the patch-level attention module.

\section{Conclusion}
\label{sec:conclusion}
We propose a full-reference image quality assessment approach called Region-Adaptive Deformable Network (RADN). 
We first revisit the classic residual blocks and propose the modified residual blocks from the viewpoint of IQA tasks, which are used to build our baseline. 
We introduce the patch-level attention mechanism for information interaction among the patch regions and the reference-oriented deformable convolution for adaptation to images of significant differences.
We also propose a contrastive pretraining strategy to further improve the model’s capability to truly distinguish the image quality rather than directly learning the regression of the quality score.
The experimental results reveal the excellent effectiveness of the proposed method. 
Our ensemble method ranked fourth in the NTIRE 2021 Perceptual Image Quality Assessment Challenge.

{\small
\bibliographystyle{ieee_fullname}
\bibliography{egbib}
}

\end{document}